\documentclass[10pt]{article} % For LaTeX2e
\usepackage[preprint]{tmlr}
\usepackage{hyperref}
\usepackage{url}
\usepackage{graphicx}
\usepackage{booktabs}
\usepackage{amsmath}

\title{Deep Feature Pyramid Convolutional Networks with In-Place Activated Batch Normalization for Automated Skin Lesion Boundary Segmentation}

\author{%
Glib Kechyn
}

\begin{document}

\maketitle

\begin{abstract}
Segmentation of skin lesion boundaries in dermoscopic imaging is an important prerequisite step for computer-aided diagnosis of malignant melanoma, but remains challenging due to fuzzy margins, occluding artifacts such as hair and blood vessels, low contrast, and high inter-patient variability. This work presents a memory-efficient deep convolutional neural network framework for lesion boundary segmentation, developed for the ISIC 2018 Challenge (Task 1: Lesion Boundary Segmentation). A U-Net-style encoder--decoder architecture is adapted using Wide ResNet38 and Dual Path Network (DPN) backbones with a Feature Pyramid Network (FPN) decoder, pretrained on ImageNet. In-Place Activated Batch Normalization (InPlace-ABN) reduces memory consumption during training, allowing higher-capacity ensembling under standard GPU memory constraints, combined with five-fold cross-validation, extensive augmentation, snapshot ensembling, and test-time augmentation. The best-performing model achieves a Thresholded Jaccard score of 0.752 on the ISIC 2018 Challenge evaluation, with single-model configurations scoring between 0.700 and 0.750; InPlace-ABN reduces memory consumption by approximately 25\%. To contextualize this result against current practice, a simplified single-model U-Net baseline retrained in 2026 using standard modern tooling is also reported, achieving a Dice score of 0.8443 (IoU 0.7608, Thresholded Jaccard 0.6680), which highlights the substantial contribution of ensembling and pretraining to the original result. This work does not claim state-of-the-art segmentation accuracy; rather, it documents a memory-efficient training approach that enables higher-capacity ensembling under constrained GPU resources, a practical consideration for training high-capacity segmentation models in settings with limited computational budgets.
\end{abstract}

\section{Introduction}
\label{sec:intro}

Malignant melanoma represents one of the most aggressive forms of skin cancer, accounting for the majority of skin-cancer-related deaths globally despite comprising a smaller proportion of overall skin tumor cases \citep{esteva2017dermatologist}. Visual inspection remains the primary frontline diagnostic technique; however, clinical evaluation alone exhibits considerable inter- and intra-observer variability and depends heavily on the experience of the dermatologist. Total body photography and digital dermoscopy are routinely used to evaluate high-risk patients, but manual delineation and semantic segmentation of lesion boundaries by pathologists remains labor-intensive, subjective, and tedious, creating a bottleneck in high-volume clinical workflows.

To address this, automated Computer-Aided Diagnosis (CAD) systems leveraging computer vision have become a central focus of biomedical image analysis. Precise lesion boundary segmentation is an essential prerequisite step in automated pipelines, since downstream algorithms calculate key clinical indicators, such as asymmetry, border irregularity, and structural color variation, directly from the segmented boundary contour. Despite substantial progress, automatic skin lesion border detection remains challenging: dermoscopic images exhibit extensive variability due to differences in skin pigmentation, fuzzy or low-contrast margins, and occluding artifacts including hair, air bubbles, ruler marks, and blood vessels.

With the advent of deep Convolutional Neural Networks (CNNs), semantic segmentation accuracy has advanced substantially across standard benchmarks \citep{codella2019isic}. Fully convolutional architectures, particularly encoder--decoder networks based on the U-Net topology \citep{ronneberger2015u}, have demonstrated strong capacity for capturing multi-scale spatial representations. More recently, the field has expanded toward hybrid Vision Transformers \citep{hatamizadeh2022swin}, foundation models such as the Segment Anything Model \citep{kirillov2023segment,ma2024segment}, and generative approaches such as diffusion models \citep{chen2023generalist} to better capture global context and refine ambiguous lesion margins.

A persistent, separate concern in deep biomedical segmentation is the tension between model capacity and hardware efficiency. Deeper CNN backbones \citep{he2016deep,zagoruyko2016wide,chen2017dual}, multi-scale decoders \citep{lin2017feature}, and high-parameter Vision Transformers improve fine-grained boundary extraction, but incur substantial GPU memory overhead during training, since standard batch normalization and activation layers store intermediate feature maps in memory for use during backpropagation. This often forces a choice between downsampled inputs and sub-optimal batch sizes. In-Place Activated Batch Normalization (InPlace-ABN) \citep{bota2018inplace} addresses this bottleneck directly by executing normalization and activation in-place.

This work builds on an ensembled U-Net-style pipeline for the ISIC 2018 Challenge introduced by \citet{kechyn2018lesion}, extending it with a systematic evaluation of the memory-efficiency contribution of InPlace-ABN and a 2026 single-model baseline comparison using standard modern tooling. The primary contribution of this work is not to establish state-of-the-art segmentation accuracy; several architectures published since 2018, including subsequent U-Net-based and attention-based models, report higher accuracy on this benchmark. Rather, this work documents a memory-efficient training approach that enabled higher-capacity ensembling under standard GPU memory constraints, and provides a transparent, present-day reference point for interpreting that original result.

\section{Related Work}
\label{sec:related}

Encoder--decoder segmentation architectures built on the U-Net topology \citep{ronneberger2015u} remain a standard approach for biomedical image segmentation. Deep residual and multi-branch backbones \citep{he2016deep,zagoruyko2016wide,chen2017dual}, combined with multi-scale decoders such as Feature Pyramid Networks \citep{lin2017feature}, have improved fine-grained segmentation performance, at the cost of increased training-time memory consumption. More recent work has explored attention-based refinements \citep{oktay2018attention}, multi-stage refinement architectures such as DoubleU-Net \citep{jha2020doubleu}, hybrid Vision Transformer architectures \citep{hatamizadeh2022swin}, and general-purpose foundation models for segmentation \citep{kirillov2023segment,ma2024segment}, as well as diffusion-based generative approaches \citep{chen2023generalist}. In-Place Activated Batch Normalization \citep{bota2018inplace} is a memory-optimization technique that reduces the memory footprint of the normalization--activation sequence common to most of these architectures, independent of the specific backbone or decoder chosen.

\section{Method}
\label{sec:method}

\subsection{Dataset and Preprocessing}
This study uses the ISIC 2018: Skin Lesion Analysis Toward Melanoma Detection challenge dataset (Task 1: Lesion Boundary Segmentation) \citep{codella2019isic,tschandl2018ham10000}. The dataset consists of dermoscopic RGB images ranging in resolution from $576\times768$ to $6748\times4499$ pixels, with binary ground-truth boundary masks. Per the official challenge structure, the training set comprised 2594 images, the validation set 100 images, and the test set 1000 images; test set ground truth was withheld by the challenge organizers and evaluated via submission-based scoring using the Thresholded Jaccard metric. Images were resized to $224\times224$ and normalized using the mean and standard deviation computed from the training set, which was further partitioned into five stratified folds to preserve class distribution during cross-validation.

\subsection{Data Augmentation}
Approximately 14 augmentation types were applied per image using the Albumentations library \citep{buslaev2018albumentations}, including motion blur, median blur, random contrast and brightness adjustment, shift-scale-rotation, contrast-limited adaptive histogram equalization (CLAHE), sharpening, elastic distortion, hue-saturation shifts, and grayscale conversion, each with randomized coefficients.

\subsection{Network Architecture}
\begin{figure}[h]
\centering
\includegraphics[width=0.6\linewidth]{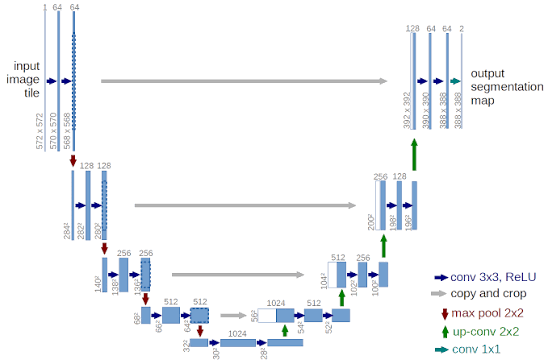}
\caption{U-Net-style encoder--decoder topology used as the base architecture.}
\label{fig:unet}
\end{figure}

Wide ResNet38 \citep{zagoruyko2016wide} and Dual Path Network (DPN) \citep{chen2017dual} backbones, pretrained on ImageNet, are used as feature-extraction encoders. The overall encoder--decoder topology is illustrated in Figure~\ref{fig:unet}.

\begin{figure}[h]
\centering
\includegraphics[width=0.6\linewidth]{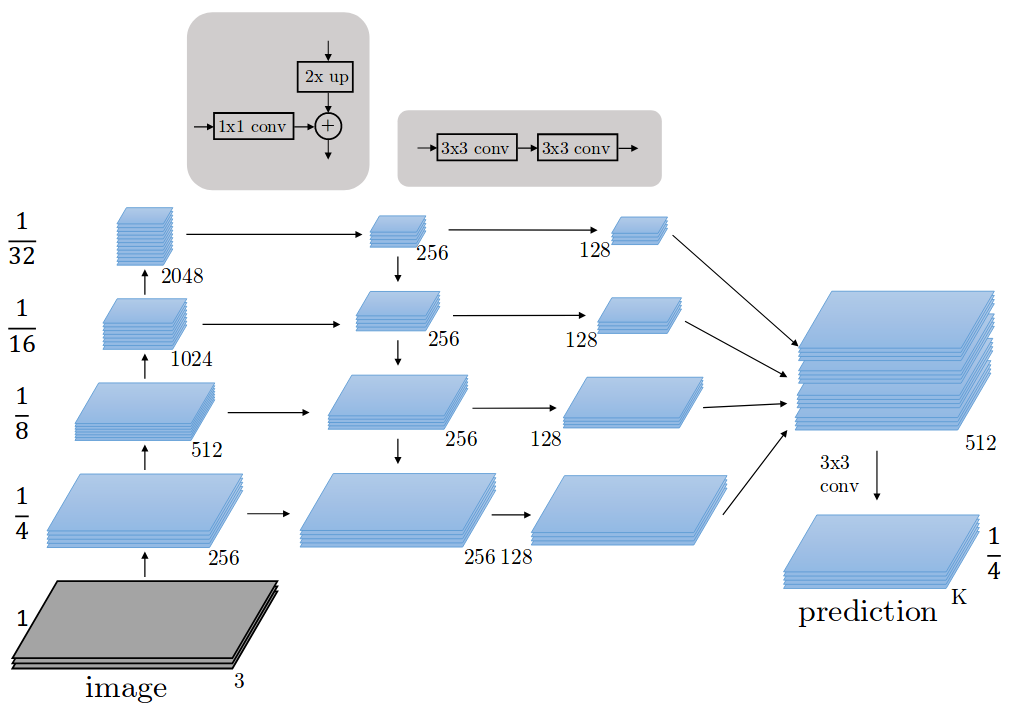}
\caption{Feature Pyramid Network (FPN) architecture used as the decoder.}
\label{fig:fpn}
\end{figure}

An FPN-based decoder \citep{lin2017feature} fuses low-resolution, high-semantic feature maps from upper encoder layers with localized feature maps from lower encoder stages via lateral skip connections, as shown in Figure~\ref{fig:fpn}. As an enhancement over the standard U-Net decoder, an initial block using a $7\times7$ convolution with stride 2, followed by max-pooling with stride 2, is applied.

\subsection{Memory-Efficient In-Place Activated Batch Normalization}
Standard deep CNN architectures execute batch normalization followed by a non-linear activation sequentially, requiring both sets of intermediate activations to be stored in GPU memory for backpropagation. InPlace-ABN \citep{bota2018inplace} merges normalization and activation into a single in-place operation, recovering forward activations during the backward pass by mathematically inverting the stored outputs rather than allocating additional memory buffers. This reduces memory consumption by approximately 25\%, enabling larger batch sizes on standard GPU hardware.

\subsection{Loss Function}
Training uses a weighted compound loss combining Binary Cross-Entropy (BCE) and soft Dice loss in equal proportion:
\begin{equation}
\mathcal{L} = 0.5 \cdot \mathcal{L}_{\text{BCE}} + 0.5 \cdot (1 - \text{Dice}).
\end{equation}
Binary cross-entropy encourages confident prediction of the lesion body, while the Dice term encourages precise boundary localization.

\subsection{Training Strategy and Post-Processing}
Models are trained using cyclic learning rate scheduling, with snapshot ensembling across multiple architectures and training snapshots, combined with hypercolumn feature aggregation \citep{hariharan2015hypercolumns}. Predicted probability maps are refined using marker-based watershed segmentation, with connected components used as markers following dilation and erosion. Final predictions are generated using test-time augmentation, averaging outputs across 7 augmented versions of each input image.

\subsection{Evaluation Metrics}
Performance is evaluated using the Thresholded Jaccard Index, the official ISIC 2018 Challenge Task 1 metric, in which the standard Jaccard (IoU) score for an image is set to zero if it falls below 0.65:
\begin{equation}
\text{IoU} = \frac{\text{TP}}{\text{TP} + \text{FP} + \text{FN}},
\end{equation}
where TP, FP, and FN denote true positives, false positives, and false negatives, respectively.

\subsection{2026 Single-Model Baseline Re-implementation}
\label{sec:baseline2026}
To provide a point of comparison using current standard tooling, a simplified single-model U-Net \citep{ronneberger2015u} was independently re-implemented and trained on the same ISIC 2018 Task 1 dataset. Unlike the pipeline described above, this baseline uses a single architecture without backbone pretraining, batch normalization, ensembling, cross-validation, test-time augmentation, or InPlace-ABN. Images and masks were resized to $256\times256$ and split 80/20 into training and validation sets. Training used the Albumentations library for augmentation (rotation up to 35 degrees, horizontal and vertical flipping), a BCEWithLogitsLoss objective, the Adam optimizer (learning rate $3\times10^{-4}$), a batch size of 32, and automatic mixed-precision training (\texttt{torch.cuda.amp}). Training ran for up to 40 epochs with early stopping based on validation Dice stagnation (patience = 10 epochs); the full 40 epochs completed without early stopping triggering.

\section{Results}
\label{sec:results}

\subsection{Segmentation Performance}
The best-performing model, an ensemble of Wide ResNet38 and DPN encoders with an FPN decoder and InPlace-ABN, achieves a Thresholded Jaccard score of 0.752 on the ISIC 2018 Challenge evaluation. Individual architecture and single-model configurations, without full ensembling, score between 0.700 and 0.750.

\subsection{Impact of In-Place Activated Batch Normalization}
Integrating InPlace-ABN reduces memory consumption by approximately 25\% relative to standard batch normalization, enabling larger batch sizes without exceeding standard GPU memory limits, and allowing higher-capacity encoder ensembles to be trained without downsampling inputs beyond the $224\times224$ working resolution.

\subsection{Comparison with a 2026 Single-Model Baseline}
\begin{figure}[h]
\centering
\includegraphics[width=0.9\linewidth]{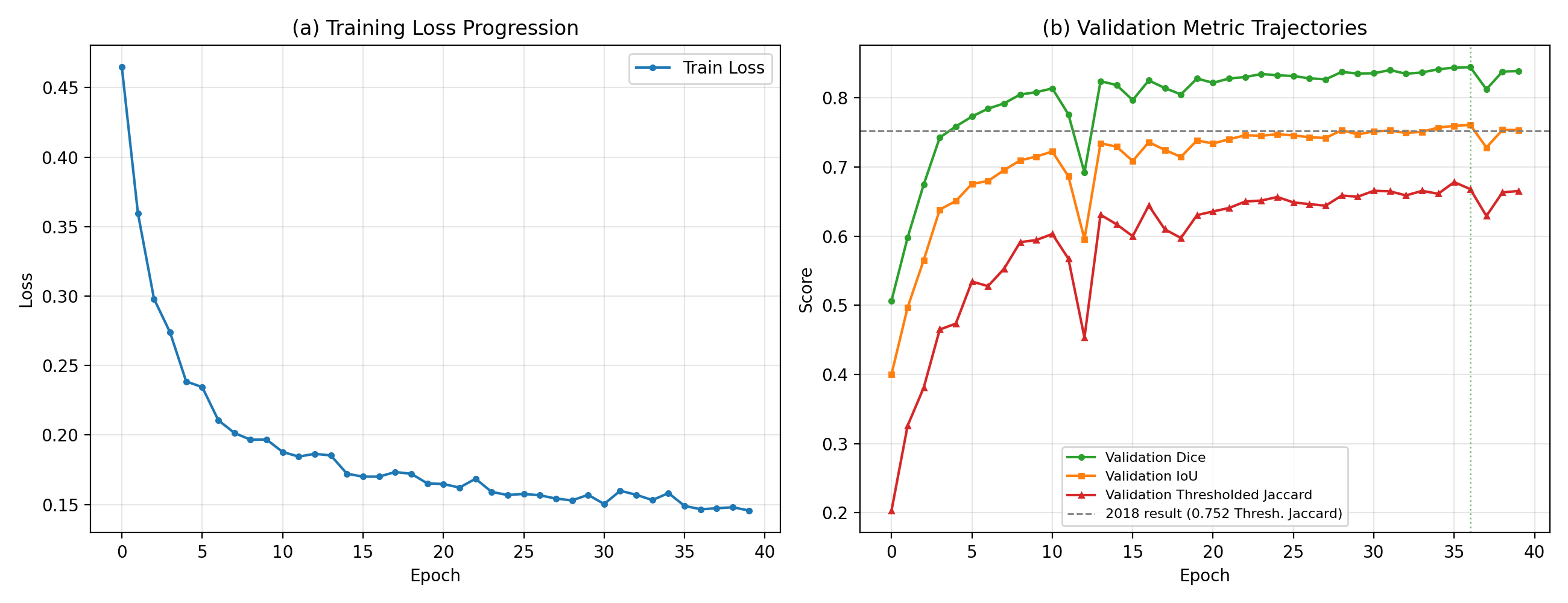}
\caption{Training and validation convergence curves for the 2026 single-model baseline over 40 epochs. (a) Training loss progression; (b) Validation Dice, IoU, and Thresholded Jaccard trajectories, with the original ensembled result (0.752 Thresholded Jaccard) shown for reference.}
\label{fig:convergence}
\end{figure}

The simplified single-model U-Net baseline described in Section~\ref{sec:baseline2026} achieves a best validation Dice score of 0.8443, corresponding to an IoU of 0.7608 and a Thresholded Jaccard score of 0.6680, after 40 training epochs. Figure~\ref{fig:convergence} shows the training loss and validation metric trajectories over the full training run. This result is lower than the 0.752 Thresholded Jaccard achieved by the ensembled pipeline, consistent with the absence of backbone pretraining, batch normalization, ensembling, and test-time augmentation in this simplified baseline. Note that the 2026 baseline and the original pipeline are not a controlled comparison in the strict sense: they differ simultaneously in backbone architecture, normalization strategy, ensembling, and augmentation scope, any of which could independently affect the reported gap. The comparison is included to give a modern, minimal-effort reference point rather than to isolate the effect of any single design choice.

\begin{figure}[h]
\centering
\includegraphics[width=\linewidth]{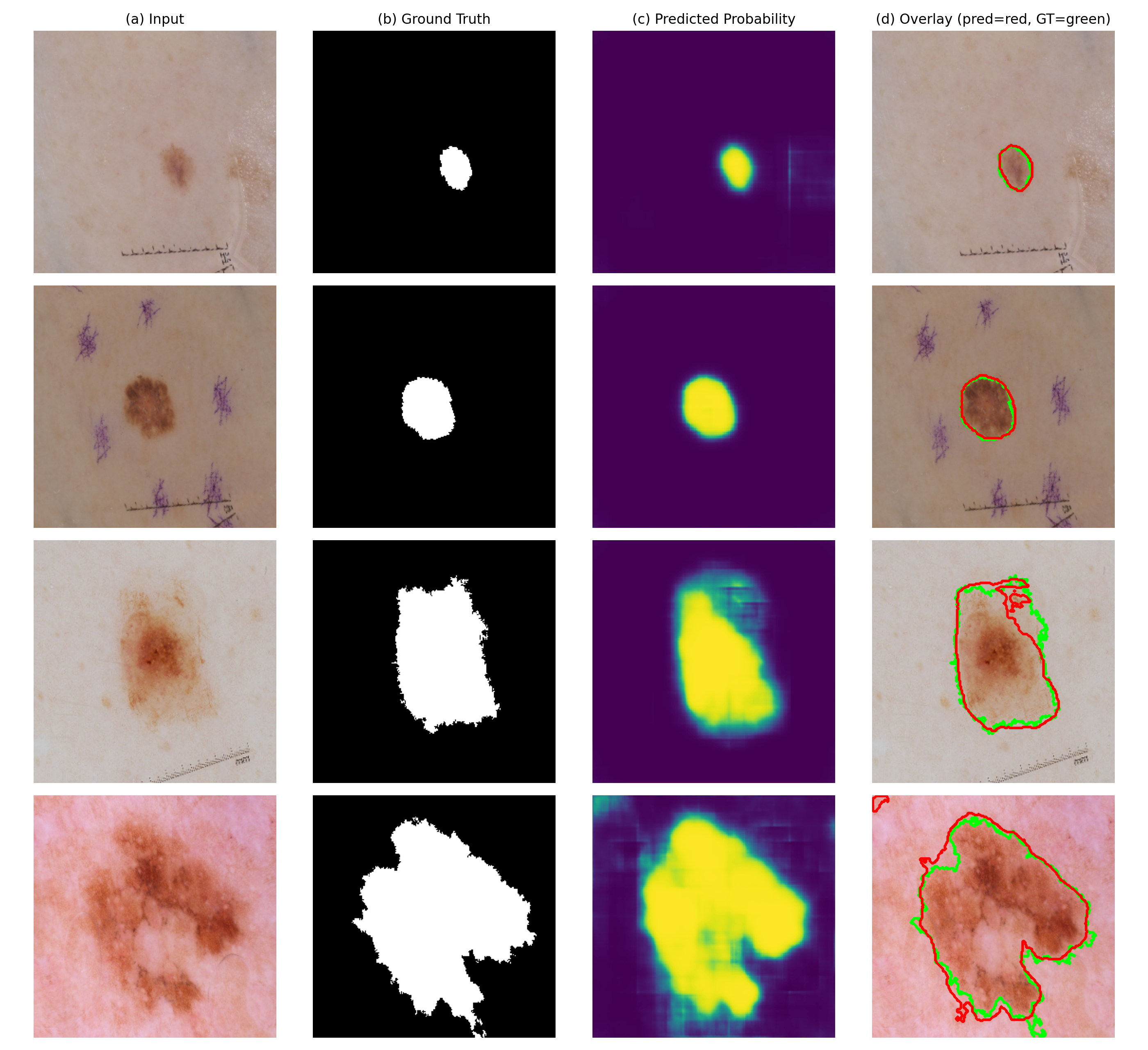}
\caption{Qualitative segmentation results from the 2026 single-model baseline on representative validation samples. From left to right: (a) input dermoscopic image, (b) ground-truth boundary mask, (c) predicted probability map, and (d) overlay of predicted boundary (red) against ground truth (green).}
\label{fig:qualitative}
\end{figure}

Figure~\ref{fig:qualitative} presents qualitative segmentation results from this baseline on representative validation samples.

\begin{table}[h]
\centering
\caption{Segmentation performance summary on the ISIC 2018 Challenge Task 1 evaluation.}
\label{tab:results}
\begin{tabular}{lccc}
\toprule
\textbf{Configuration} & \textbf{Dice} & \textbf{IoU} & \textbf{Thresholded Jaccard} \\
\midrule
Ensembled (Wide ResNet38 + DPN, FPN, InPlace-ABN) & -- & -- & 0.752 \\
Single-model configurations (no ensembling) & -- & -- & 0.700--0.750 \\
2026 simplified single-model U-Net baseline & 0.8443 & 0.7608 & 0.6680 \\
\bottomrule
\end{tabular}
\end{table}

\section{Discussion}
\label{sec:discussion}

This work documents a memory-efficient ensembling approach to automated skin lesion boundary segmentation. By integrating InPlace-ABN into an ensemble of Wide ResNet38 and DPN encoders with an FPN decoder, the approach achieves a Thresholded Jaccard score of 0.752, within the range reported by the broader field of ISIC 2018 challenge participants, though below the top-ranked submissions, which reached approximately 0.802 \citep{codella2019isic}. Notably, top-performing challenge submissions already approached previously reported inter-observer agreement between human annotators (Jaccard $\approx 0.786$) \citep{codella2019isic}, underscoring the difficulty of the task even for expert annotators.

The 2026 single-model baseline provides additional context for interpreting the original result: despite using modern default tooling and automatic mixed-precision training, the absence of ensembling, backbone pretraining, and test-time augmentation results in a meaningfully lower Thresholded Jaccard score (0.668 vs.\ 0.752). This suggests that these techniques, rather than the choice of normalization strategy alone, account for a substantial portion of the original pipeline's performance.

\subsection{Limitations and Future Work}
Several factors likely limited the achieved accuracy relative to top-performing submissions. Post-competition architectural refinements, including attention mechanisms \citep{oktay2018attention}, multi-stage refinement approaches such as DoubleU-Net \citep{jha2020doubleu}, and more recently transformer-based and foundation-model approaches \citep{hatamizadeh2022swin,kirillov2023segment}, have since demonstrated improved boundary precision on this benchmark. Additionally, working at $224\times224$ resolution, chosen to accommodate GPU memory constraints available at the time, likely limited fine-grained boundary precision relative to approaches using higher-resolution inputs. Future extensions of this work could incorporate these more recent architectural advances alongside the memory-efficient training approach demonstrated here, and could re-evaluate the accuracy/memory tradeoff at higher input resolutions now that InPlace-ABN-style memory optimizations are more widely available in standard deep learning frameworks.

\section{Conclusion}
\label{sec:conclusion}

This work presents a memory-efficient ensemble segmentation pipeline for automated skin lesion boundary detection. By integrating In-Place Activated Batch Normalization with Wide ResNet38 and DPN encoder backbones and an FPN decoder, the approach achieves a Thresholded Jaccard score of 0.752 while reducing GPU memory consumption by approximately 25\% relative to standard batch normalization. A simplified single-model baseline retrained in 2026 using standard modern tooling achieves a lower Thresholded Jaccard score of 0.668, highlighting the substantial contribution of ensembling, pretraining, and test-time augmentation to the original result. This memory efficiency enables larger-batch training of high-capacity ensembles under standard GPU hardware constraints, illustrating the practical value of memory-efficient normalization techniques for biomedical image segmentation under limited computational resources.

\subsubsection*{Broader Impact Statement}
This work concerns a preprocessing step (lesion boundary segmentation) within computer-aided diagnosis pipelines for skin cancer screening, not a standalone diagnostic tool. As with any machine-learning component intended for eventual clinical adjacency, outputs of this kind should be used to assist, not replace, clinical judgment, and any deployment in a diagnostic pipeline would require appropriate clinical validation, regulatory review, and evaluation for performance disparities across skin tones and lesion types, which is not assessed in this work. All experiments use a publicly available, de-identified benchmark dataset (ISIC Archive); no new patient data was collected.

\bibliography{tmlr_references}
\bibliographystyle{tmlr}

\end{document}